\documentclass[10pt,twocolumn,letterpaper]{article}

\usepackage{cvpr}
\usepackage{times}
\usepackage{epsfig}
\usepackage{graphicx}
\usepackage{amsmath}
\usepackage{amssymb}
\usepackage{mmstyle}
\usepackage{subcaption}
\usepackage{color}
\usepackage{booktabs}
\usepackage{pifont}

\usepackage[pagebackref=true,breaklinks=true,letterpaper=true,colorlinks,bookmarks=false]{hyperref}

\cvprfinalcopy %

\def\cvprPaperID{5679} %

\ifcvprfinal\fi
\begin{document}

\title{Intra- and Inter-Action Understanding via Temporal Action Parsing}

\author{Dian Shao \quad Yue Zhao \quad Bo Dai \quad Dahua Lin \\
	CUHK-SenseTime Joint Lab, The Chinese University of Hong Kong\\
	{\tt\small \{sd017, zy317, bdai, dhlin\}@ie.cuhk.edu.hk} 	
}

\maketitle
\thispagestyle{empty}

\setlength{\abovedisplayskip}{3pt}
\setlength{\belowdisplayskip}{3pt}

\begin{abstract}
Current methods for action recognition primarily rely on deep
convolutional networks to derive feature embeddings of visual and
motion features. While these methods have demonstrated remarkable performance
on standard benchmarks, we are still in need of a better understanding
as to how the videos, in particular their internal structures, relate to
high-level semantics,
which may lead to benefits in multiple aspects, \eg~interpretable predictions
and even new methods that can take the recognition performances to a next level.
Towards this goal, we construct TAPOS, a new dataset developed on sport videos
with manual annotations of sub-actions, and conduct a study on temporal
action parsing on top~\footnote{Project website: \url{https://sdolivia.github.io/TAPOS/}}.
Our study shows that a sport activity usually consists of multiple sub-actions
and that the awareness of such temporal structures is beneficial to action
recognition.
We also investigate a number of temporal parsing methods,
and thereon devise an improved method that is capable of mining 
sub-actions from training data without knowing the labels of them.
On the constructed TAPOS, 
the proposed method is shown to reveal \textbf{intra}-action information,
~\ie how action instances are made of sub-actions, 
and \textbf{inter}-action information,
~\ie one specific sub-action may commonly appear in various actions.
\end{abstract}

\section{Introduction}
\label{sec:intro}

Action understanding is a central topic in computer vision,
which benefits a number of real-world applications,
including video captioning~\cite{xiong2018move}, video retrieval~\cite{hu2011survey,shao2018find} and vision-based robotics~\cite{mataric2002sensory}.
Although over the past decade, remarkable progress has been made on 
action classification~\cite{yang2020tpn, wang2013action,tran2015learning, feichtenhofer2016convolutional},
action localization~\cite{zhao2017temporal,lin2018bsn},
and action segmentation~\cite{lea2016segmental,Kuehne_2016,ding2018weakly},
the insight into actions themselves remains lacking,
as few works have analyzed actions at a finer-granularity,
such as exploring their internal structures (intra-action understanding),
discovering their mutual relationships (inter-action understanding).

One hindrance for intra- and inter-action understanding is a well-annotated dataset,
which provides annotations that penetrate into actions,
besides action labels as in most existing datasets~\cite{soomro2012ucf101,kuehne2011hmdb,karpathy2014large,kay2017kinetics}.
However,
such a dataset is hard to collect,
especially when labeling sub-actions.
Specifically,
action labels provided by humans are sometimes ambiguous and inconsistent,
\eg~\emph{open/close fridge} are treated as the same action while \emph{pour milk/oil} belong to different actions.
Such an issue could become severer when we deal with sub-actions,
as compared to actions, sub-actions share more subtle differences between each other.
Moreover,
sub-actions belonging to the same category could not only appear in different instances of some action,
but also instances of different actions.
Although previous attempts have alleviated these issues 
by restricting the types of both actions and sub-actions within relatively more formatted cases,
\eg~instructional~\cite{tang2019coin,miech2019howto100m} and cooking~\cite{rohrbach2012database,kuehne2014language,stein2013combining,fathi2011learning} videos.
In more general cases,
ensuring a consistent labeling scheme across sub-actions may be infeasible,
considering the scale of a dataset.

Fortunately,
we observe that humans are sensitive to \emph{boundaries} of sub-actions,
even without knowing their categories.
We thus provide intra-action annotations in the form of high-quality temporal segmentations,
instead of sub-action labels.
The temporal segmentations divide actions into segments of different sub-actions,
implicitly revealing actions' internal structures.
The constructed dataset,
which we refer to as 
\textbf{Temporal Action Parsing of Olympics Sports (TAPOS)},
contains over $16K$ action instances in $21$ Olympics sport classes.
We focus on Olympics sport actions as they have consistent and clean backgrounds,
and diverse internal structures and rich sub-actions.
These characteristics would encourage models to exploit the action themselves rather than the background scenes.

On top of TAPOS,
we notice a temporal segmental network (TSN)~\cite{wang2016temporal} can obtain significant performance gains when the segments are aligned with temporal structures,
instead of being evenly divided.
Motivated by the study,
we propose to investigate actions by temporally parse an action instance into segments,
each of which covers a complete sub-action,
where categories of these sub-actions are unknown.
\eg~parsing an instance of \emph{triple jump} into six segments,
whose semantics could be characterized as \emph{run-up}, three \emph{jumps}, and then a \emph{reset}.
While conceptually simple, temporal action parsing (TAP) is challenging in several aspects.
Firstly,
there are no pre-defined sub-action classes,
and the associations among segments, \ie~which segments belong to the same class,
are also unknown.
Consequently,
the possible number of distinct classes could be as large as $N * M$ ($> 30k$ in TAPOS), 
where $N$ is the number of action instances and $M$ is the average number of segments in an instance.
This characteristic of TAP highlights its difference with tasks having pre-defined classes such as \emph{Action Segmentation}~\cite{rohrbach2012database,stein2013combining,fathi2011learning,gao2014jhu},
since it is infeasible to turn TAP into these tasks by enumerating over possible class assignments.
Moreover,
compared to action boundaries,
at the finer granularity of sub-actions,
the transition between consecutive segments is often quite smoother,
making it difficult to localize their boundaries.

We further develop an improved framework for temporal action parsing on TAPOS,
inspired by recently proposed Transformer~\cite{vaswani2017attention}.
The proposed framework, referred to as TransParser,
adopts two stacked transformer as its core,
where frames of action instances are used as the queries,
and parameters in a memory bank are served as keys and values.
While TransParser outperforms baselines on temporal action parsing,
its structure also enables it to discover semantic similarities of sub-action segments within one action class and across different action classes,
in an unsupervised way.
By investigating TransParser,
we could also reveal additional intra-action information (\eg~which sub-action is the most discriminative one for some action class)
and inter-action information (\eg~which sub-action commonly appears in different action classes).

The contribution of this work can be briefly summarized into
three aspects:
1) a new dataset TAPOS which provides a class label for each action instance
as well as its temporal structure;
2) a new task, namely Temporal Action Parsing, that encourages the exploration
of the internal structures of actions;
3) an improved framework for temporal action parsing,
which provides additional abilities for further intra- and inter-action understanding.

\section{Related Work}
\label{sec:related}

\textbf{Datasets.} Being an important task in computer vision,
various datasets have been collected for action understanding,
which could be roughly divided into three categories.
Datasets in the first category provides only class labels,
including early attempts (\eg~KTH~\cite{laptev2004recognizing}, Weizmann~\cite{blank2005actions}, UCFSports~\cite{rodriguez2008action}, Olympic~\cite{niebles2010modeling}) of limited scale and diversity,
and succeeding benchmarks (\eg~UCF101~\cite{soomro2012ucf101}, HMDB51~\cite{kuehne2011hmdb}, Sports1M~\cite{karpathy2014large}, and Kinetics~\cite{kay2017kinetics})
that better fit the need of deep learning methods.
However,
despite of increasing numbers of action instances being covered,
more sophisticated annotations are not provided by these datasets.
In the second category, datasets provide boundaries of actions in untrimmed videos.
Specifically,
videos in THUMOS'15~\cite{THUMOS15}~contain action instances of $20$ sport classes.
And daily activities are included in ActivityNet~\cite{caba2015activitynet}~and Charades~\cite{sigurdsson2016hollywood}.
Other datasets in this category include HACS~\cite{zhao2019hacs} and AVA~\cite{gu2018ava}.
Although these datasets are all annotated with temporal boundaries,
they focus on the location of an action in an untrimmed video.
Instead, we intend to provide boundaries inside action instances,
revealing their internal structures.

Our proposed dataset belongs to the third category,
where fine-grained annotations for action instances are provided.
Most of the existing datasets in this category focus on instructional videos,
such as cooking videos in 50 Salads~\cite{stein2013combining}, Breakfast~\cite{kuehne2014language}, and MPIICooking~\cite{rohrbach2012database},
as well as surgical videos in JIGSAWS \cite{gao2014jhu}.
Compared to these datasets,
TAPOS mainly focuses on instances of Olympics sport actions for two reasons.
First,
Olympics actions have rich sub-actions,
and loosely formatted but diverse internal structures,
so that models are encouraged to exploit inside actions in a data-driven way.
Moreover,
instances of the same Olympics action have consistent and clean backgrounds,
making models focus on the action itself.

\vspace{2pt}
\textbf{Tasks.} %
Various methods have been proposed for vision-based action recognition~\cite{yang2020tpn,shao2020finegym,wang2013action,oneata2013action,simonyan2014two,tran2015learning,feichtenhofer2016convolutional,carreira2017quo,varol2018long,feichtenhofer2018slowfast,dai2017detecting,fang2018pairwise,li2019transferable},
where they are asked to predict a single class label for a given video instance.
Temporal action localization~\cite{xu2017r,zhao2017temporal,lin2018bsn}, 
on the other hand,
aims at
identifying temporal locations of action instances in an untrimmed video.
Another line of research focuses on a detailed understanding of the internal structures of action instances, especially along the temporal dimension.
Specifically, in the task of action recognition, some researchers \cite{pirsiavash2014parsing,wang2013latent} 
implicitly learn the temporal structures of complex activities to promote performances,
but the quality of estimated temporal structures is not assessed. 
In temporal action parsing (TAP), we emphasize the importance of such temporal structures, 
and provide annotations for quality assessment.
The most related task to TAP is temporal action segmentation (TAS)~\cite{lea2017temporal,kuehne2014language,alayrac2017joint,lei2018temporal,Farha_2019_CVPR,richard2018action,li2019weakly}.
TAS aims at labeling each frame of an action instance within a set of pre-defined sub-actions, 
which can be done in a fully-supervised~\cite{Farha_2019_CVPR,lea2016segmental}~(\eg frame labels are provided) 
or a weakly-supervised~\cite{li2019weakly,richard2018action,richard2017weakly}~(\eg only ordered sub-actions are provided) manner.
While TAS relies on a pre-defined set of sub-actions, assuming all samples contain only these classes, 
TAP offers only the boundaries between sub-actions, which are significantly weaker supervisions.
We empirically found in our experiments that methods for TAS cannot well estimate the temporal structures under the setting of TAP,
indicating TAP poses new challenges.

\begin{figure*}[ht]
	\centering
	\includegraphics[width=1.0\linewidth]{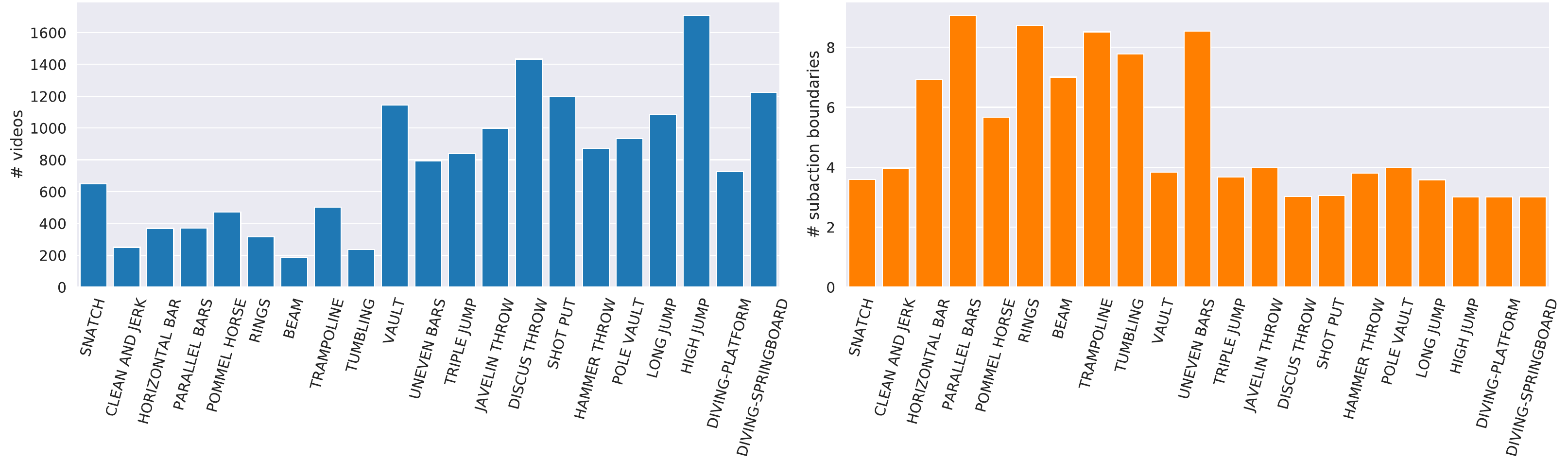}
	\caption{\small Statistics of the dataset: the left histogram depicts the average number of timestamps per class; the right one illustrates the average number of annotations for sub-action boundaries.}
	\label{fig:data_stat}
	\vspace{-10pt}
\end{figure*}

\section{Dataset}
To encourage intra- and inter-action understanding,
we construct a new dataset,
referred to as Temporal Action Parsing of Olympics Sports (TAPOS).
Specifically,
samples in TAPOS are all action instances of Olympics sports,
so that instances belonging to the same sport tend to have a consistent background.
Moreover, 
samples in TAPOS are ensured to cover a complete action instance with no shot changes.
These two characteristics of TAPOS make it a suitable dataset for models that focus on the action itself,
as potential distracters are explicitly avoided.
For each sample,
we provide annotations of two levels of granularity,
namely the action labels (\eg~\emph{triple-jump}, \emph{shot-put}, etc),
and the ranges of sub-actions (\eg~\emph{run-up}, \emph{jump} and \emph{landing} in \emph{triple-jump}),
in the form of temporal timesteps.
It's worth noting that labels of sub-actions are not provided.
While sub-actions such as \emph{run-up} could be further decomposed into stages at a finer granularity,
in this paper we restrict our annotations to have only two-level granularities,
leaving finer annotations as future work. 
We start by introducing the collection process,
followed by dataset statistics and dataset analysis.

\subsection{Dataset Collection}

To obtain samples in TAPOS,
we at first collect a set of videos from public resources (\eg~Youtube).
Each collected video will be divided into a set of shots utilizing techniques for shot detection~\cite{apostolidis2014fast} to obtain instances within a single shot.
For action labels and ranges of sub-actions,
we apply a two-round annotation process using crowdsourcing services.
In the first round,
irrelevant shots and shots containing incomplete action instances are filtered out,
and remaining shots are labeled with action classes.
Subsequently, every remaining shot will be assigned to three annotators, so that we could cross-validate the annotations.
Each annotator will mark the boundaries of consecutive sub-actions independently.
Before the second round,
we will provide annotators with instructional descriptions and illustrative samples,
guiding them to divide shots without knowing sub-action labels.
Finally, we filter out shots with a number of temporal timestamps less than $3$.

\subsection{Dataset Statistics}

\begin{figure}
	\centering
		\includegraphics[width=0.9\linewidth]{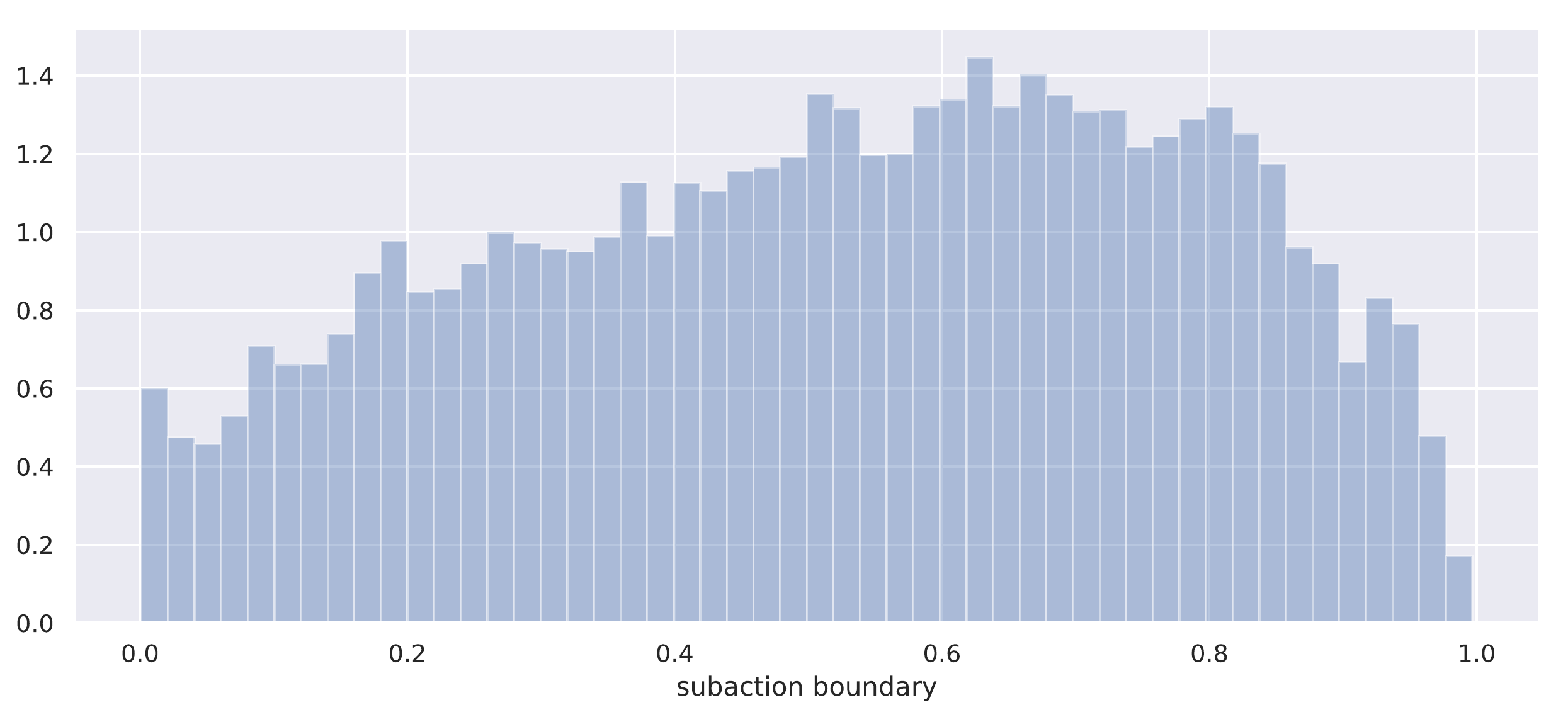}
		
	\caption{\small Probability distribution of sub-action boundary occurrence across the video. Length of each video is normed to one.}
	\label{fig:data_prob}
	\vspace{-10pt}
\end{figure}
\begin{figure*}
	\centering
		\includegraphics[width=0.8\linewidth]{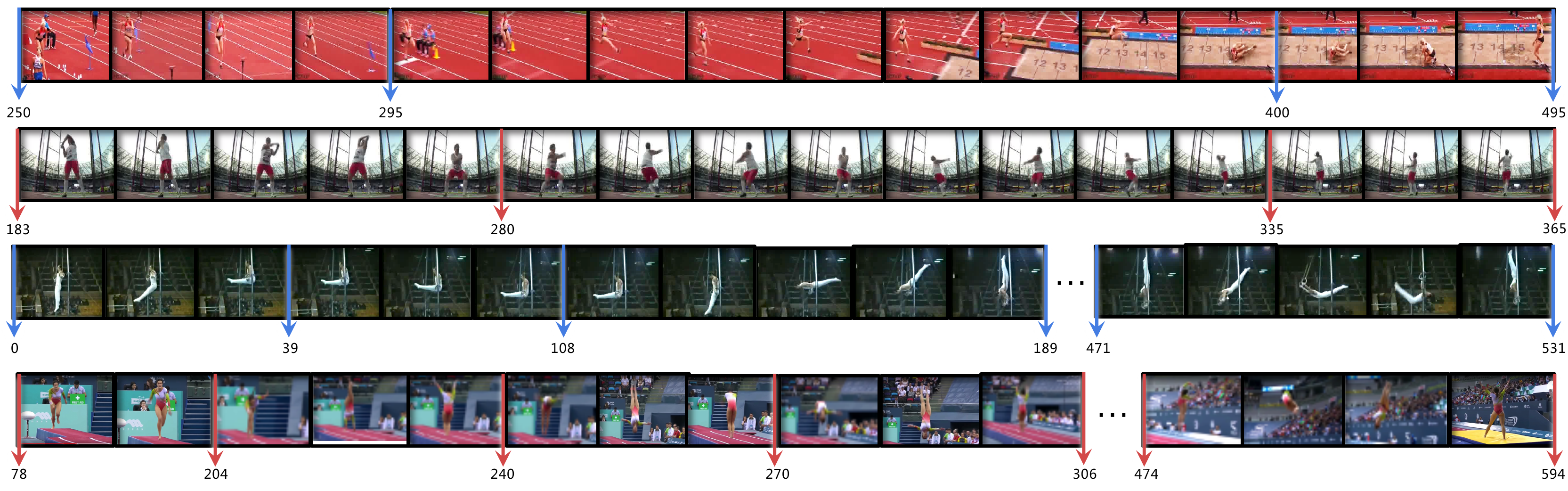}
	\caption{\small Some samples from the proposed TAPOS dataset. From
top to bottom, action classes are: triple jump, hammer throw, rings and tumbling. Temporal boundaries of sub-actions are annotated, complex actions are therefore composed of temporally adjacent subactions, \eg, hammer throw in the second row comprises \emph{swing}, \emph{rotate boby} and \emph{throw}. }
	\label{fig:data_overview}
	\vspace{-10pt}
\end{figure*}

TAPOS contains $16,294$ valid instances in total,
across $21$ action classes.
These instances have a duration of $9.4$ seconds on average.
The number of instances within each class is different,
where the largest class \emph{high jump} has over $1,600$ instances,
and the smallest class \emph{beam} has $200$ instances.
The average number of sub-actions also varies from class to class,
where \emph{parallel bars} has $9$ sub-actions on average,
and \emph{long jump} has $3$ sub-actions on average, as shown in Figure~\ref{fig:data_stat}.
Finally, as Figure~\ref{fig:data_prob}~ shows,
start and end points of sub-actions could temporally be any way for a single instance.
While the number of instances within each class reflects the natural distribution of action classes,
the variance in instances including their time durations, numbers of sub-actions in them and locations of sub-actions has reflected the natural diversity of actions' inner structures,
facilitating more sophisticated investigations on actions.

All instances are split into train, validation and test sets,
of sizes 13094, 1790, and 1763, respectively.
When splitting instances,
we ensure that instances belonging to the same video will appear only in one split.

\section{Analysis of Sub-actions}

\begin{table}[t]
\centering
\small
\setlength{\tabcolsep}{0.5pt}
\begin{tabular}{cp{3mm} cp{3mm}c p{3mm} cp{3mm}c}
\toprule
TSN      && \multicolumn{3}{c}{RGB}    && \multicolumn{3}{c}{RGB + Flow}   \\ \cmidrule{1-1} \cmidrule{3-5} \cmidrule{7-9}
sample            && Top-1 Acc. && Avg. Acc. && Top-1 Acc. && Avg. Acc. \\ \cmidrule{1-1} \cmidrule{3-3} \cmidrule{5-5} \cmidrule{7-7} \cmidrule{9-9}
uniform           && 83.97      && 82.22     && 91.01      && 88.15          \\ 
aligned           && 88.83      && 86.22          && 93.80      && 91.73          \\ \bottomrule
\end{tabular}
\caption{\small Comparison of performance on action classification using different sampling for TSN. Both the top-1 accuracy and overall accuracy is reported. }
\label{tab:tsn}
\end{table}

\begin{figure}
	\centering
		\includegraphics[width=0.83\linewidth]{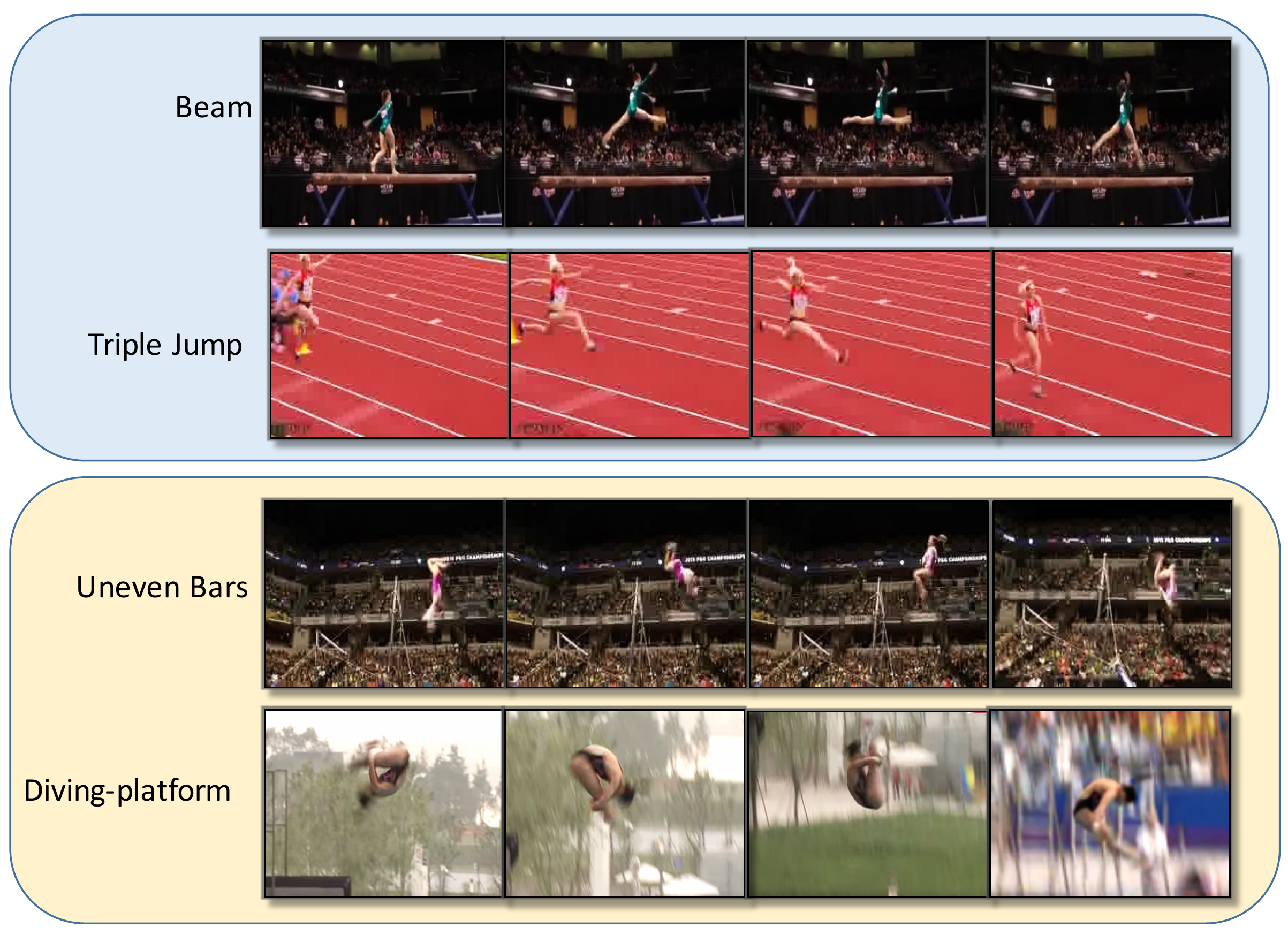}
		
	\caption{\small Similar sub-actions are shared by irrelevant actions, \eg, jump in \emph{beam} and \emph{triple jump} (the first pair), somersault in \emph{uneven bars} and \emph{diving} (the second pair). }
	\label{fig:data_patterns}
	\vspace{-10pt}
\end{figure}

The success of feature-based methods, \ie~those that directly
connect visual feature embeddings to action classes, leads to a question:
\emph{``do we need to take a step further into the temporal structures?''}.
In this section we present a brief study for this question.

Indeed, the utility of temporal structures have been demonstrated,
sometimes in an implicit way, in previous works.
Specifically,
Wang \etal~\cite{wang2016temporal} found that dividing a video
into segments helps action recognition. However, this work only considers
even segmentation that is not necessarily aligned with the inherent structure.
Feichtenhofer \etal~\cite{feichtenhofer2018have}
observed that certain patterns emerge automatically from the internal activations of a model for action classification.
These discoveries indicate that each class of actions often contain
temporal structures in certain ways.
This corroborates with our intuition that an action is often composed of
stages at finer granularity, \eg~the entire process of a \emph{long jump}
consists of a \emph{run-up}, a \emph{jump}, and a \emph{landing}.
We refer to segments of an action as \emph{sub-actions}.

Next, we further investigate how the decomposition of an action into
sub-actions influence action understanding.
In the first study, we compared temporal segmental networks~\cite{wang2016temporal}
on TAPOS, with two configurations: (1) with segments of even durations, and
(2) with segments aligned with annotated sub-actions.
Table~\ref{tab:tsn} shows that the latter configuration outperforms the former
by a large margin, which implies that the use of temporal structures, in particular
the segmentation of sub-actions, can significantly help the performance
of action recognition.
In the second study, we carefully examine the connections between sub-actions
in different action classes.
As shown in Figure~\ref{fig:data_patterns},
sub-actions in different action classes can be similar, even for those actions
that appear to be quite different in the first place.
These findings suggest that to effectively discriminate between such
classes, one may need to go beyond a local scope and resort to a more
global view, \eg~looking into how sub-actions evolve from one to another.

\begin{figure*}[t]
	\centering
	\small
	\includegraphics[width=0.9\linewidth]{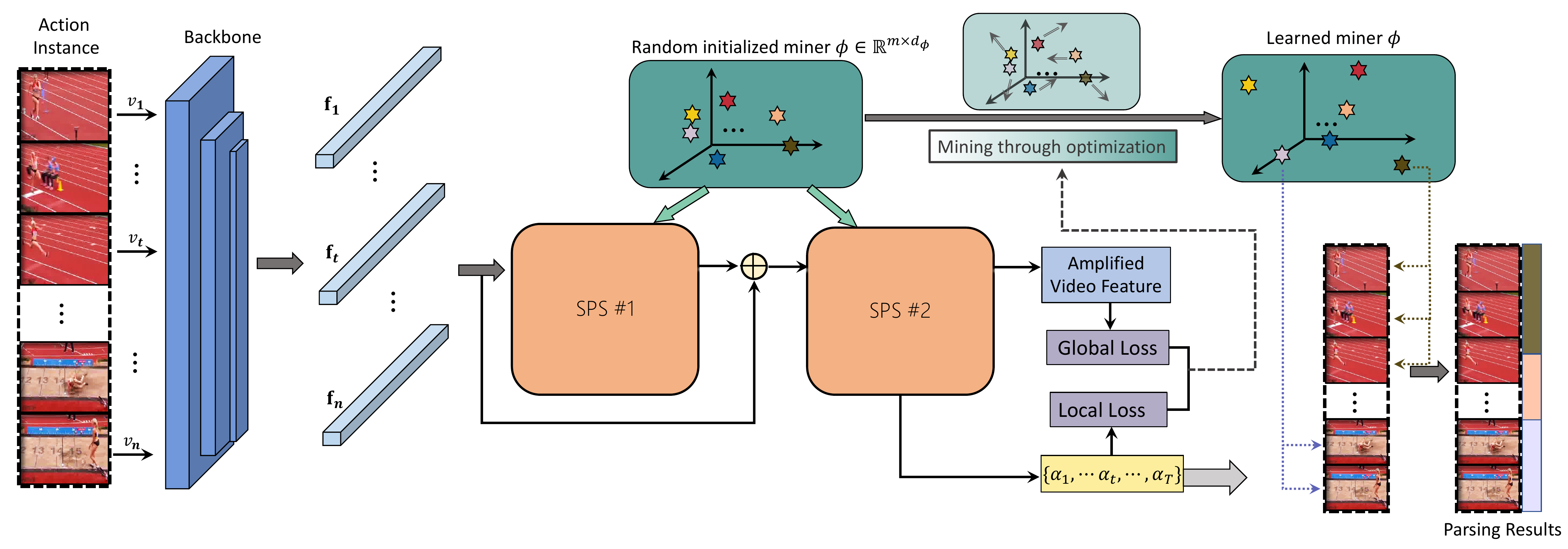}
	\caption{\small An overview of the proposed TransParser. Given a sequence of video frames, we first obtain the frame-level feature $ \{\vf_1, \cdots, \vf_n \} $.
	Lying in the core of the TransParser are two stacked Soft-Pattern-Strengthen (SPS) Units which maintain a pattern miner $\phi$ and use a soft-attention mechanism to produce amplified feature from $ \vf_t $.
	We use two losses, \ie a local loss to promote agreement between frames within a sub-action while suppressing that across sub-actions and a global loss to predict action label as a regularization.
	Throughout optimization, different representations for different sub-actions are automatically learned in the pattern miner.
	During inference, the calculated attention weight at the last SPS Unit can be used to obtain the temporal action parsing results.}
	\label{fig:transparser}
	\vspace{-10pt}
\end{figure*}

\begin{figure}[t]
	\centering
	\small
	\includegraphics[width=0.8\linewidth]{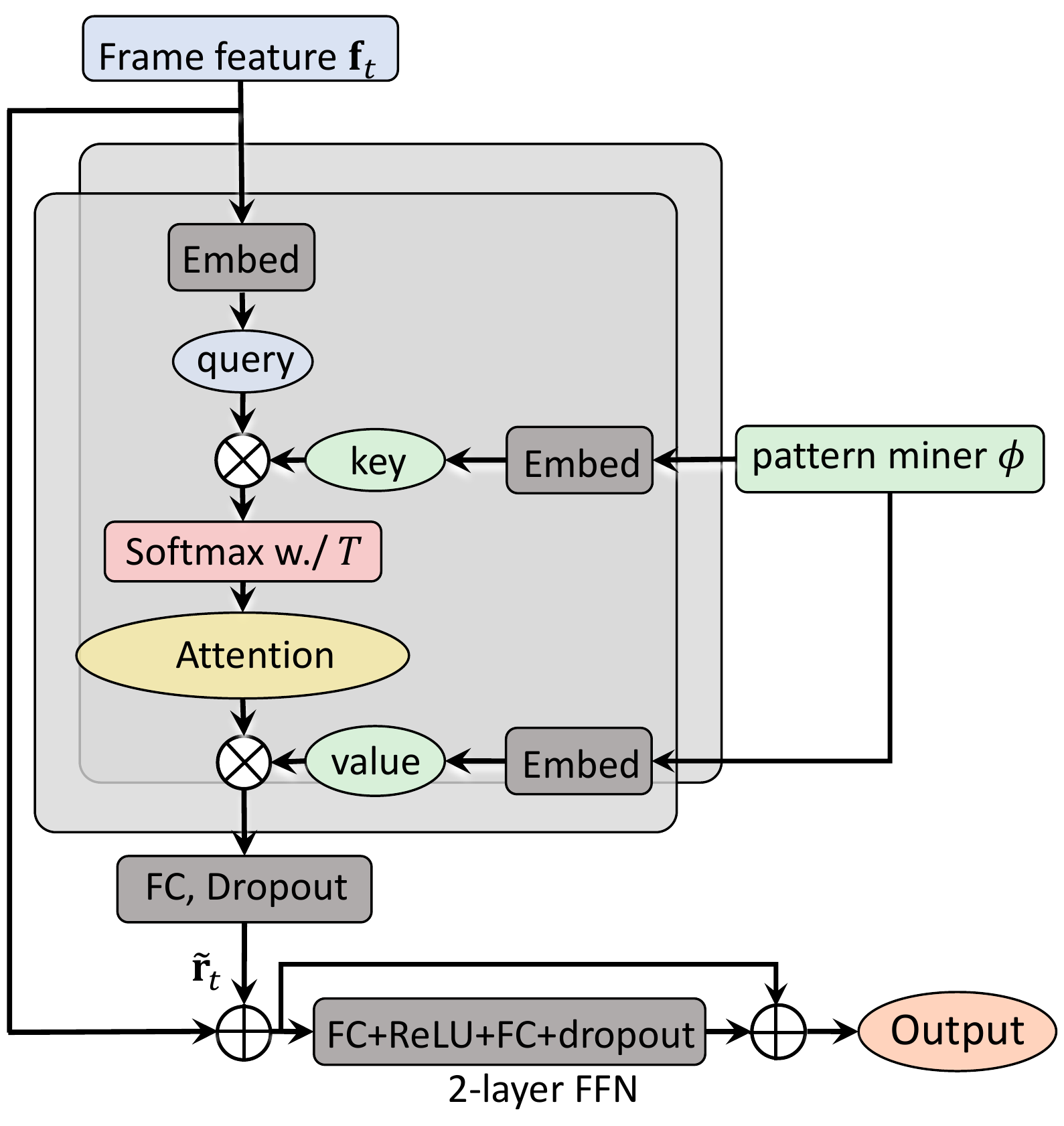}
	\caption{\small An instantiation of SPS Unit.
	It maintains a pattern miner $\phi$ and takes frame feature $ \vf_t $ as input.
	A multi-head soft-attention is conducted thereon.
	The final output $ \widetilde{\vr}_t $ is added to $ \vf_t $ to achieve amplification.}
	\label{fig:sps}
	\vspace{-10pt}
\end{figure}

\section{Temporal Action Parsing}
\label{sec:method}
In this section, we first briefly introduce the setting of temporal action parsing,
and then discuss our framework. %

\subsection{Task Definition}
Formally, let $A = \{v_1, ..., v_n \}$ denotes a certain action instance of $n$ frames,
and $S_1, ..., S_k$ be its corresponding sub-actions
so that $A = \{S_1, ..., S_k\}$, where $S_i = \{ v_{t_i}, ..., v_{t_{i+1} - 1}\}$.
$A$ can then be represented by a set of middle-level representations,
each derived from one of its sub-action.
The goal of an action parsing model is thus to identify the starting frames $\{v_{t_1}, ..., v_{t_k}\}$ of sub-actions.

\subsection{TransParser for Temporal Action Parsing}

To decompose an action instance into a set of sub-actions without knowing the possible sub-action categories,
we develop a data-driven way to discover the distinct patterns of different sub-actions,
as shown in Fig.~\ref{fig:transparser}.
Specifically,
given frames of an action instance $\{v_1, ..., v_n\}$,
we at first apply a BNInception network~\cite{ioffe2015batch} to extract per-frame features $\{\vf_1, ..., \vf_n; \vf_t \in \Rbb^{1 \times d_f}\}$.
Each feature $\vf_t$ is then refined by a Soft-Pattern-Strengthen (SPS) unit.
The SPS unit maintains a parametric pattern miner $\vphi$ to learn distinct characteristics of sub-actions,
which could be used to regularize the input feature, amplifying its discriminative %
patterns.
The refinement could be described as $\vf'_t = \vf_t + \mathrm{SPS}(\vf_t, \vphi)$.

The parsing process of TransParser works as follows.
Given refined features of action frames $\{\vf'_1, ..., \vf'_n\}$,
we at first compute the responses $\{\valpha_1, ..., \valpha_n\}$ of them and the patterns stored in $\vphi$.
The representative pattern (\ie~$\argmax_j \valpha_t|_j$) in each response is selected thereafter,
and once two consecutive frames $(t, t+1)$ have different representatives, TransParser marks the start of a new sub-action at $(t+1)$-th frame.

\vspace{-10pt}
\paragraph{Soft-Pattern-Strengthen (SPS) Unit.}
Some components of the SPS unit are inspired by Transformer~\cite{vaswani2017attention},
which has been actively studied in the language domain but rarely explored in the field of action understanding.
Being designed to amplify the discriminative patterns in the input feature $\vf_t$,
the SPS unit will maintain a pattern miner, parameterized as $\vphi = [\vphi_1, ..., \vphi_m]$ of size $m \times d_\phi$,
to discover informative patterns of frame features during training.
Such a process is conducted by a soft-attention that treats the input feature $\vf_t$ and the miner $\vphi$ respectively as query and (key, value) pairs:
\begin{align}
	 \valpha_t & = \mathrm{softmax}((\vf_t \cdot \mW_Q) \cdot (\vphi \cdot \mW_K)^T), \label{eq:response} \\
	 \vr_t & = \valpha_t \cdot (\vphi \cdot \mW_V), 
\end{align} 
where $\cdot$ stands for matrix multiplication, and the output $\vr_t$ can be intuitively regarded as a residual of $\vf_t$ helpful to distinguish subtle differences.
Following~\cite{vaswani2017attention} in practice we use multi-head attention consisting of two groups of $\{\mW_Q, \mW_K, \mW_V\}$,
and the final $\widetilde{\vr}_t$ is obtained by feeding the two features $\vr^{(1)}_t$ and $\vr^{(2)}_t$ from each group respectively into a single fc layer.
The output feature $\vf'_t$ is computed by $\mathrm{FFN}(\vf_t + \widetilde{\vr}_t)$,
where FFN stands for a small feed-forward net as in~\cite{vaswani2017attention}.
Amplification is achieved via adding $\widetilde{\vr}_t$ to $\vf_t$.

We utilize the combination of two losses to learn the TransParser, with the ground-truths of temporal segmentations and labels of action instances.
1) \emph{local loss}: 
to help the pattern miner $\vphi$ capture informative patterns in features of action frames,
a semantic loss is applied to maximize the agreement between frames within a sub-action while suppressing that across sub-actions:
\begin{align}
	\cL_\mathrm{local} & = \frac{\cL_\mathrm{sim} + \lambda}{\cL_\mathrm{dissim}}, \\
	\cL_\mathrm{sim} & = \mathrm{avg}(\sum_{t_1, t_2 \in S_i \forall i} || \valpha'_{t_1} - \valpha'_{t_2} ||_2), \\
	\cL_\mathrm{dissim} & = \mathrm{avg}(\sum_{t_1 \in S_i, t_2 \in S_j, i \ne j} || \valpha'_{t_1} - \valpha'_{t_2}||_2),
\end{align}
where $\valpha'_t$ is the response computed on $\vf'_t$ according to Eq.\eqref{eq:response},
as the refined feature $\vf'_t$ contains amplified discriminative patterns compared to $\vf_t$.
$\lambda$ is a regularizer to prevent trivial solutions (\eg~all $\valpha$s are collapsed to be the same one hot vector).
2) \emph{global loss}:
We further add a global classification loss as a regularization,
suggesting that refined features of action frames still need to be representative to action categories.
For each action instance $A$ with $n$ frames:
\begin{align}
	\cL_\mathrm{global} & = \mathrm{NLL}( \frac{1}{n}\sum_{t=1}^n (\mW \cdot \vf'_t), l_A),
\end{align}
where $\mW$ is the weight of a classifier, and $l_A$ is the label.
For conciseness, in practice we apply a second SPS unit to obtain $\valpha'_t$ for the local loss.
Succeedingly, we also use $\vf''_t$ from the second unit in the global loss, which is more discriminative than $\vf'_t$.

\section{Experiments}
\subsection{Evaluation Metrics}
Once the parsing process is finished, %
we can obtain a series of predicted start frames denoted by $ \cT_\cP = \{ s_1, s_2, \cdots, s_{\vert \cP \vert} \} $.
Assuming the ground-truth sub-actions start at $ \cT_\cG = \{ t_1, t_2, \cdots, t_{\vert \cG \vert} \} $, we can determine the correctness of each prediction if its distance from the nearest ground-truth is smaller than a certain threshold $ d $.
$ d $ can either be in absolute frame number $ \Delta t $ or relative percentage $ \Delta t / T $.
The number of correct predictions is written as $ \vert \cT_\cP \widetilde{\cap}_d \cT_\cG \vert $, where $ \widetilde{\cap}_d $ can be regarded as an operation of intersection with respect to certain tolerance.
We report the recall, precision, and F1 score, defined by:
\begin{align}
\text{Recall}@d &= \frac{\vert \cT_\cP \widetilde{\cap}_d \cT_\cG \vert}{\vert \cT_\cG \vert},
\hspace{20pt}
\text{Prec}@d = \frac{\vert \cT_\cP \widetilde{\cap}_d \cT_\cG \vert}{\vert \cT_\cP \vert}, \\
\text{F1}@d &= \frac{2 \times \text{Recall}@d \times \text{Prec}@d}{\text{Recall}@d + \text{Prec}@d}
\end{align}

\subsection{Baseline methods}
Due to the connections between temporal action parsing and other tasks, such as temporal action segmentation~\cite{ding2018weakly,huang2016connectionist,lea2017temporal} and action detection~\cite{lin2018bsn}, we select representative methods from these tasks and adapt them to temporal action parsing for comparison with several modifications.

\noindent\textbf{Action boundary detection.}
We resort to a sequence model, temporal convolution network (TCN)~\cite{lea2017temporal,lin2018bsn} particularly, to estimate the emerge of action state changes.
Given a snippet of $ T $ frames, a two-layer temporal convolution network is constructed on top to densely predict a scalar for every frame. 
Following~\cite{lin2018bsn}, the annotated temporal boundary along with its $ k $ neighboring frames is labeled as 1 and the rest %
are set to be 0.
The network is optimized using a weighted Binary Cross Entropy loss due to the imbalance between positive (\ie sub-action change point) and negative samples.
During inference, the sub-action is detected once the output is over a certain threshold $ \theta_c $, \eg 0.5.

\begin{figure*}[!t]
	\centering
	\vspace{-10pt}
	\includegraphics[width=0.9\linewidth]{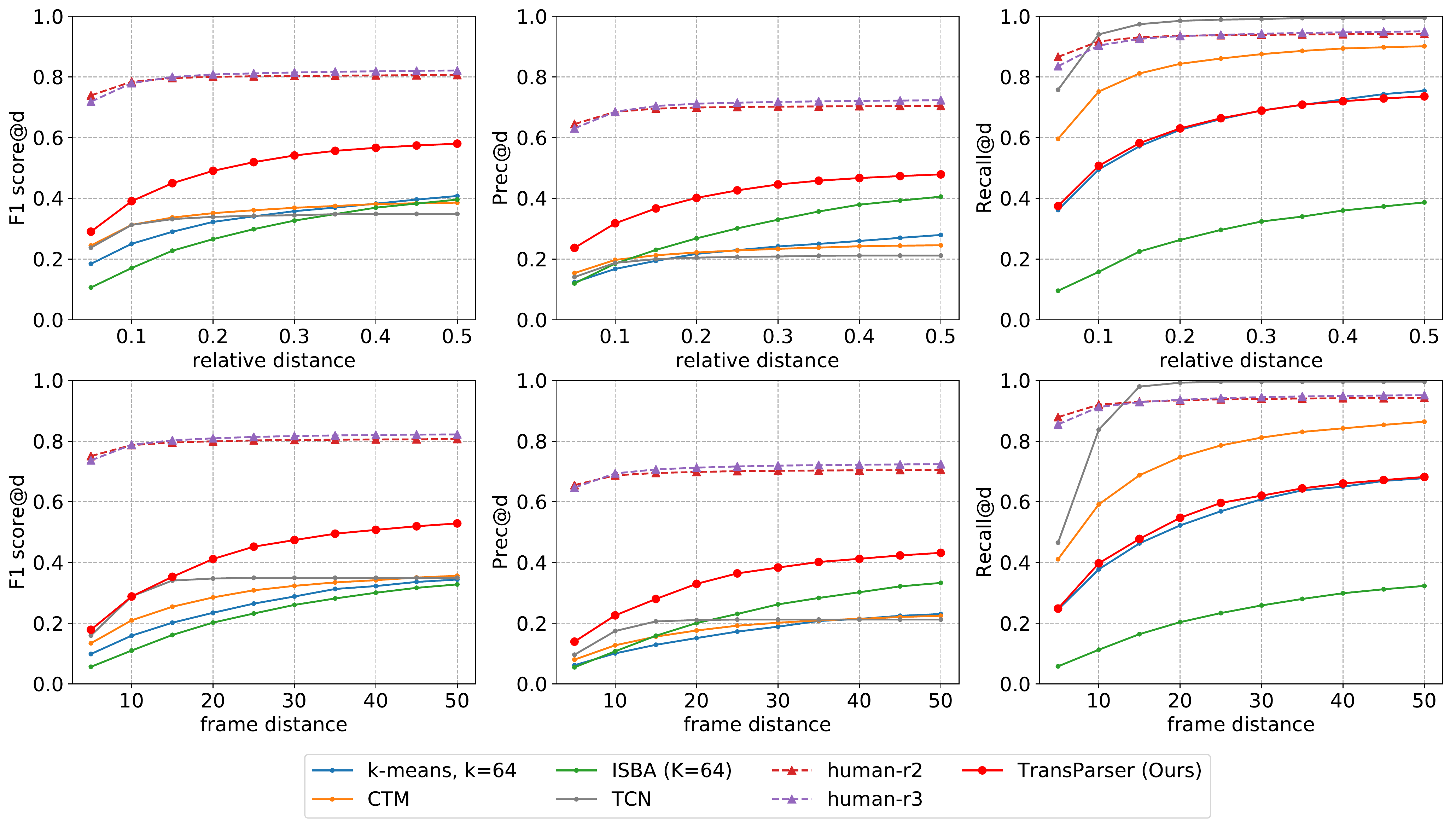}
	\caption{\small Comparison of different methods as well as human performance in terms of F1-score, precision and recall at different tolerance levels of relative distance and frame distance.}
	\label{fig:exp_quant}
	\vspace{-15pt}
\end{figure*}

\noindent\textbf{Weakly-supervised temporal action segmentation.}
Temporal action segmentation aims at labeling each frame of an action instance with a set of pre-defined sub-actions.
In the weakly-supervised setting, only a list of sub-actions in the order of occurrence without precise temporal location is provided.
We select two representative methods via Iterative Soft Boundary Assignment (ISBA)~\cite{ding2018weakly} and Connectionist Temporal Modeling (CTM)~\cite{huang2016connectionist}.
For ISBA, we generate pseudo-labels by extracting frame-level features $ \{ \vf_{i} \}_{i=1}^N $ and pre-grouping them into $ K $ clusters.
For CTM, the original training objective is to maximize the log likelihoods of the \emph{pre-defined} target labeling.
In our case, the loss is changed to the sum of log likelihoods for all \emph{possible} labelings, in that all $ k $ distinctive randomly sampled sub-actions could be a possible solution.
During inference, we use the simple best path decoding, \ie concatenating the most active outputs at every timestamp.

\subsection{Quantitative results}
\noindent \textbf{Parsing results of different methods.}
We use the aforementioned three metrics, namely $ \text{Recall}@d $, $ \text{Prec}@d $, and $ \text{F1}@d $ to evaluate the parsing performance of different methods.
We vary the relative distance from 0.05 to 0.5 at a step of 0.05 and the absolute frame distance from 5 to 50 at a step of 5.
The average F1-score across different distance thresholds is reported in Table~\ref{tab:performance}.
We can see that our method outperforms all baseline methods by a large margin.
We also calculate the overlap of the second- and third-round annotation against the first-round to be the \emph{human} performance.
We see that 1) high consistency exists between human annotators;
2) there is still a huge gap compared with human performance, leaving great room for improvement.

A detailed comparison of F1 score, precision and recall are shown in Figure~\ref{fig:exp_quant}.
As we can see, (1) Both TCN and CTM methods have exceedingly high recall but low precision, showing that these methods suffer from severe over-parsing, indicating that they focus too much on local difference; However, ISBA performs poorly on recalls but yield higher precision than CTM and TCN, indicating that such a coarse-to-fine manner may be trapped and cannot exploit intra-action information.
(2) The performance of our method consistently increases when relaxing the distance threshold, while the baseline methods quickly saturate.

\begin{table}[!t]
	\centering
	\small
	\setlength\tabcolsep{2pt}
	\begin{tabular}{c|c|c}
		\hline
		& avg. F1-score (rel.) & avg. F1-score (abs.) \\ \hline\hline
		k-means  ($ k=64 $)			& 0.3302              & 0.2881              \\ \hline
		ISBA ($ k=64 $)				& 0.2892			  & 0.2604				\\ \hline
		CTM         				& 0.3502              & 0.3194              \\ \hline
		TCN         				& 0.3303              & 0.3483              \\ \hline
		\textbf{TransParser} 		& \textbf{0.4745}     & \textbf{0.3981}     \\ \hline\hline
		Human (r2)  & 0.7948              & 0.8031              \\ \hline
		Human (r3)  & 0.8012              & 0.8158              \\ \hline
	\end{tabular}
	\caption{\small Temporal action parsing results on the proposed TAPOS dataset measured by average F1-score.}
	\label{tab:performance}
	\vspace{-10pt}
\end{table}

\begin{table}[!t]
	\centering
	\small
	\setlength\tabcolsep{4pt}
	\begin{tabular}{c|c||c|c|c}
		\hline
		\# of SPS & local loss  & avg. F1 & avg. Recall & avg. Precision \\ \hline
		$ \times 1 $ &			 & 0.2897    & 0.6950  & 0.1832   \\ \hline
		$ \times 1 $ & \ding{51} & 0.3996    & 0.5354  & 0.3189   \\ \hline
		$ \times 2 $ & \ding{51} & 0.4210    & 0.5548  & 0.3393   \\ \hline
	\end{tabular}
	\caption{\small Temporal action parsing results of TransParser under different settings.
	The average F1, recall and precision are calculated across $ d \in \{5, 10, \cdots, 50\}. $}
	\label{tab:ablation}
	\vspace{-15pt}
\end{table}

\begin{figure*}[!t]
	\vspace{-10pt}
	\centering
	\includegraphics[width=0.8\linewidth]{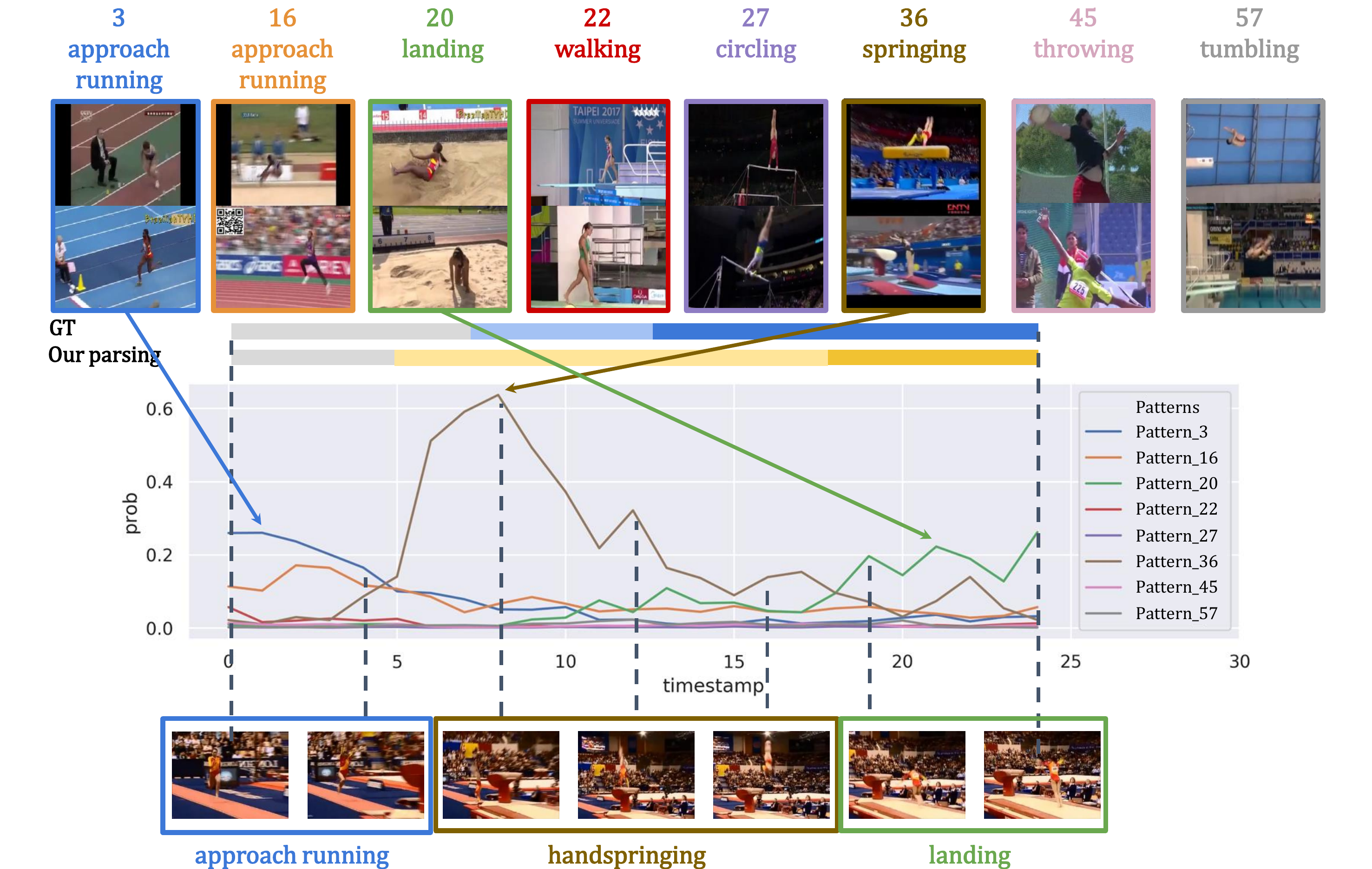}
	\caption{\small The semantic meaning of selected rows from the pattern miner is illustrated on top.
		 Given a video of a vault (bottom), we visualize the response of activated rows in the miner and provide the predicted parsing results. The figure is best viewed in color.}
	\label{fig:actom_mining}
	\vspace{-10pt}
\end{figure*}

\noindent \textbf{Variants of the proposed TransParser.}
In this part, we validate the effectiveness of the designs behind TransParser.
The results are summarized in Table~\ref{tab:ablation}.
If the local loss is dropped, we observe an increase of recall at the cost of a significant decrease in precision.
This reveals the crucial role of local semantic loss to encourage consistency between intra-stage frames and suppress that between inter-stage frames.
We can also see that increasing the number of SPS Units improves the performance, showing that discriminative differences can be amplified.
Increasing the number of SPS Units to over 2 does not yield further improvements.

\noindent \textbf{TransParser-based sampling benefits action recognition.}
We train a TSN by sampling frames based on the parsing results from Table~\ref{tab:performance}.
we can see from Table~\ref{tab:parsing_tsn} that compared to other baseline methods the parsing results by TransParser can benefit action classification by a notable improvement ($\sim 1\% $) over the uniform sampling strategy.

\begin{table}[t]
\centering
\small
\setlength{\tabcolsep}{0.5pt}
\begin{tabular}{cp{2mm} cp{2mm}c p{2mm} cp{2mm}c}
\toprule
TSN      && \multicolumn{3}{c}{RGB}    && \multicolumn{3}{c}{RGB + Flow}   \\ \cmidrule{1-1} \cmidrule{3-5} \cmidrule{7-9}
sample            && Top-1 Acc. && Avg. Acc. && Top-1 Acc. && Avg. Acc. \\ \cmidrule{1-1} \cmidrule{3-3} \cmidrule{5-5} \cmidrule{7-7} \cmidrule{9-9}
uniform           && 83.97      && 82.22     && 91.01      && 88.15     \\ 
ISBA		  && 80.95      && 79.61     && 88.88      && 85.80  		\\
CTM           && 82.51      && 82.33     && 89.83      && 88.11         \\ 
TCN			  && 81.79      && 81.10     && 90.00      && 87.16 		\\
\textbf{TransParser} && \textbf{84.80} && \textbf{83.30} && \textbf{91.62} && \textbf{89.26}          \\ \bottomrule
\end{tabular}
\caption{\small Performances of TSN \cite{wang2016temporal} on action classification using different sampling schemes.} %
\label{tab:parsing_tsn}
\vspace{-10pt}
\end{table}

\subsection{Qualitative results}

In this part, we present some qualitative analysis to gain better knowledge of TransParser.
First, results on TAPOS are shown in Figure~\ref{fig:actom_mining}.
Particularly, we retrieve video frames with highest attention score $ \alpha_{t,k} $ with respect to each row of the miner, \ie $ \phi_k $.
It is interesting to observe that different rows of the pattern miner $ \phi $ in SPS are responsive for different sub-actions, %
\eg approach running, landing, and tumbling. 
Note that both $ \phi_3 $ and $ \phi_{16} $ are most responsive to approach running but are visually different: the former is from long jump/triple jump and resembles sprinting; the latter is from high jump and %
is more similar to take-off.
Further, %
certain sub-action occurs in various actions, \eg 
the sub-action of throwing is common for both discus throwing and javelin throwing.
Finally, given a complete instance of vaulting, three stages of approach running, springing onto the vault and landing are clearly observed.

\begin{figure}
	\centering
		\includegraphics[width=0.9\linewidth]{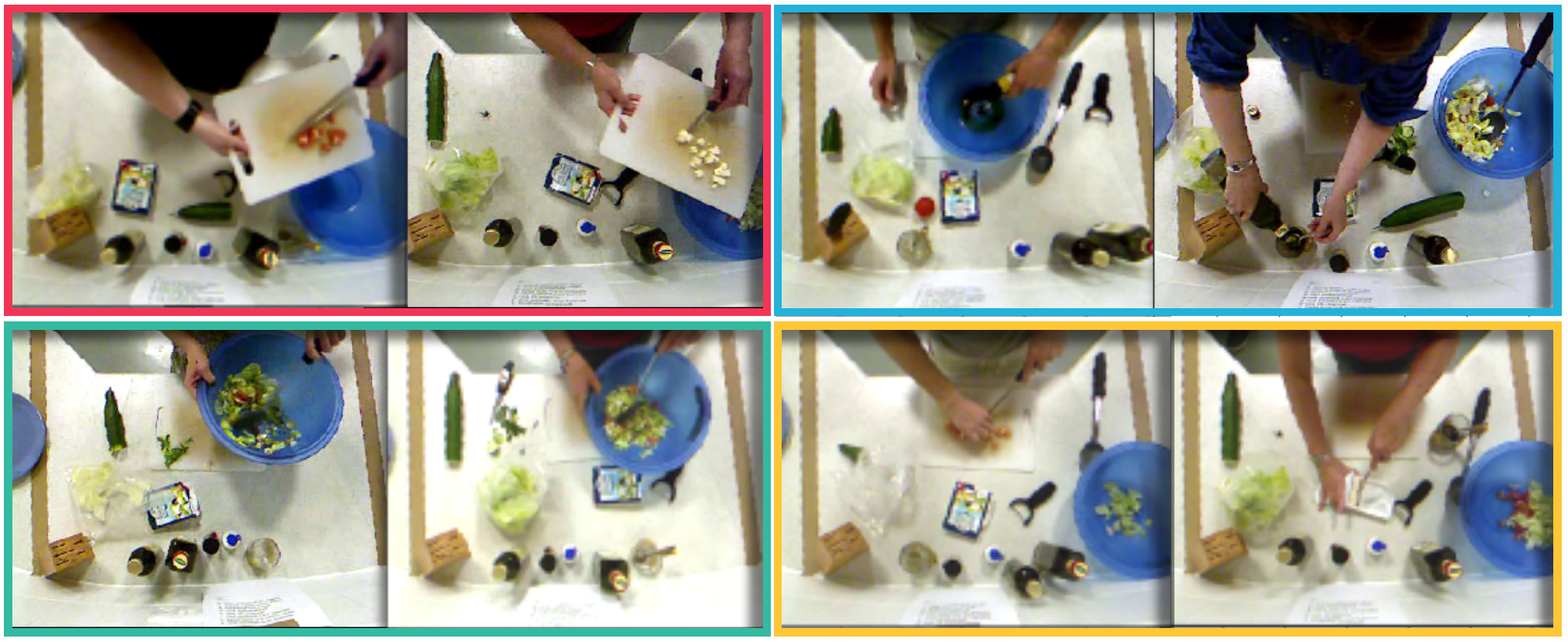}
		
	\caption{\small Qualitative analysis on 50Salads~\cite{stein2013combining}. 
Patterns from top-left to bottom-right: \emph{place tomato/cheese into bowl}, \emph{add pepper/oil}, \emph{mix dressing/ingredients}, and \emph{cut lettuce/cheese}. 
Instances in the same box are annotated to be different actions in 50 Salads, but share similar motion patterns as predicted by the TransParser.}
	\label{fig:salads}
	\vspace{-10pt}
\end{figure}

In addition, we also include a qualitative analysis on 50Salads~\cite{stein2013combining} dataset, which is commonly used for action segmentation. 
The results are shown in Figure~\ref{fig:salads}.
As we can see, the automatically mined patterns demonstrate different semantic meanings with human annotated sub-actions,
which tend to focus on motion dynamics.
For example, \emph{add pepper} and \emph{add oil} are labeled as different classes in 50Salads
while they actually follow a similar motion pattern, as predicted by the TransParser.

\vspace{-5pt}
\section{Conclusion}
\label{sec:concls}

\vspace{-5pt}
In this paper we propose a new dataset called TAPOS,
that digs into the internal structures of action instances,
to encourage the exploration towards the hierarchical nature of human actions.
In TAPOS, we provide each instance with not only a class label,
but also a high-quality temporal parsing annotation at the granularity of sub-actions,
which is found to be beneficial for sophisticated action understanding.
We also propose an improved method, TransParser, for action parsing,
which is capable of identifying underlying patterns of sub-actions without knowing the categorical labels.
On TAPOS,
TransParser outperforms existing parsing methods significantly.
Moreover, 
with the help of automatically identified patterns,
TransParser successfully reveals the internal structure of action instances,
and the connections of different action categories.

\vspace{-15pt}
\small{
\paragraph{Acknowledgements.}
This work is partially supported by SenseTime Collaborative Grant on Large-scale Multi-modality Analysis and the General Research Funds (GRF) of Hong Kong (No. 14203518
and No. 14205719).

{\small
\bibliographystyle{ieee_fullname}
\bibliography{actionparsing}
}

\end{document}